\documentclass[runningheads]{llncs}

\usepackage{eccv}

\usepackage{eccvabbrv}

\usepackage{graphicx}
\usepackage{subcaption}
\usepackage{booktabs}

\usepackage[accsupp]{axessibility}  

\newcommand{\bftab}{\fontseries{b}\selectfont}

\usepackage{hyperref}

\usepackage{orcidlink}

\begin{document}

\title{Fine-tuning a Multiple Instance Learning Feature Extractor with Masked Context Modelling and Knowledge Distillation} 

\titlerunning{Masked Context Modelling with Knowledge Distillation}

\author{Juan I. Pisula\inst{1,2} \and
Katarzyna Bozek\inst{1,2,3}}

\authorrunning{J.I. Pisula and K. Bozek}

\institute{Institute for Biomedical Informatics, Faculty of Medicine and University Hospital Cologne, University of Cologne, Germany \and
Center for Molecular Medicine Cologne (CMMC), Faculty of Medicine and University Hospital Cologne, University of Cologne, Germany \and
Cologne Excellence Cluster on Cellular Stress Responses in Aging- Associated Diseases (CECAD), University of Cologne, Germany \\
\email{juan.pisula@uk-koeln.de, k.bozek@uni-koeln.de}}







\maketitle

\begin{abstract}

The first step in Multiple Instance Learning (MIL) algorithms for Whole Slide Image (WSI) classification consists of tiling the input image into smaller patches and computing their feature vectors produced by a pre-trained feature extractor model. Feature extractor models that were pre-trained with supervision on ImageNet have proven to transfer well to this domain, however, this pre-training task does not take into account that visual information in neighboring patches is highly correlated. Based on this observation, we propose to increase downstream MIL classification by fine-tuning the feature extractor model using \textit{Masked Context Modelling with Knowledge Distillation}. In this task, the feature extractor model is fine-tuned by predicting masked patches in a bigger context window. Since reconstructing the input image would require a powerful image generation model, and our goal is not to generate realistically looking image patches, we predict instead the feature vectors produced by a larger teacher network. A single epoch of the proposed task suffices to increase the downstream performance of the feature-extractor model when used in a MIL scenario, even capable of outperforming the downstream performance of the teacher model, while being considerably smaller and requiring a fraction of its compute.

\end{abstract}

\begin{figure}[t]
  \centering
  \begin{subfigure}[c]{0.45\linewidth}
    \centering
    \includegraphics[width=.95\linewidth]{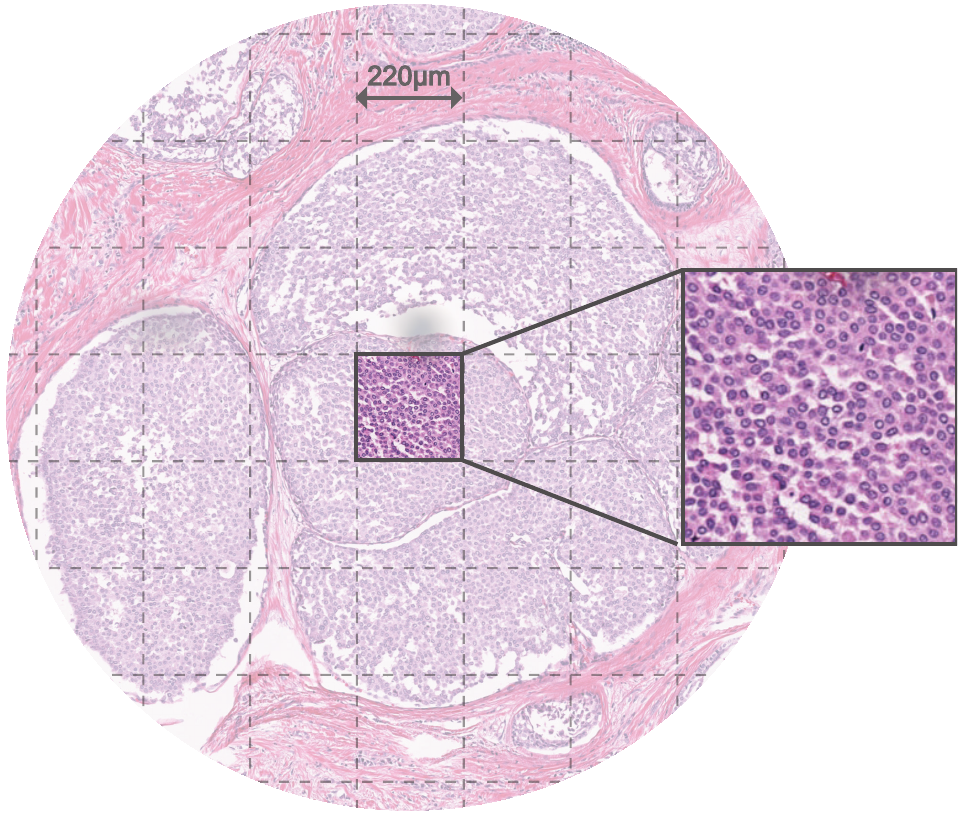}
    \caption{}
    \label{fig:1-1}
  \end{subfigure}
  \begin{subfigure}[c]{0.4\linewidth}
    \centering
    \includegraphics[width=.9\linewidth]{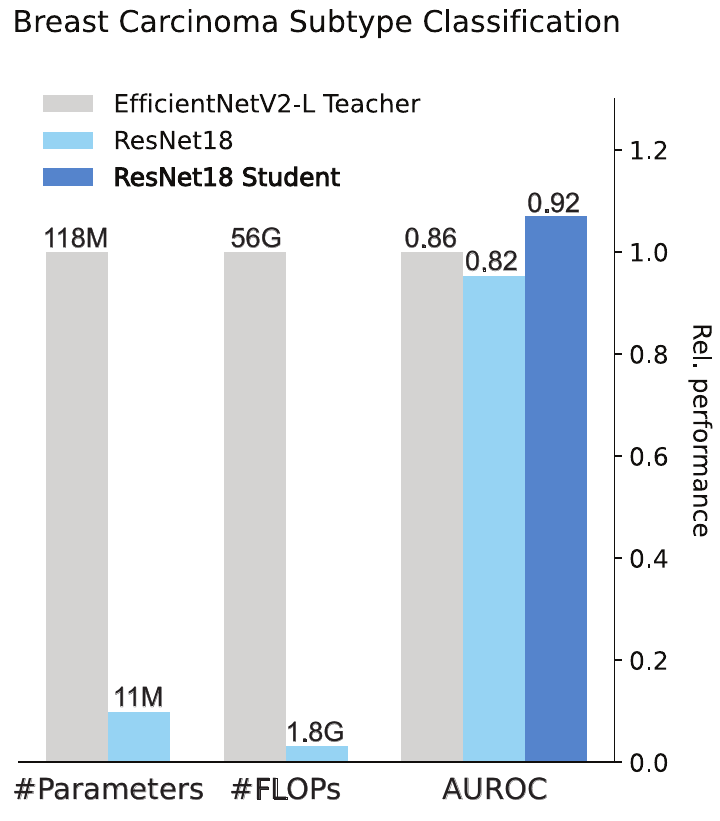}
    \vfill
    \caption{}
    \label{fig:1-2}
  \end{subfigure}
  \caption{a) A cutout of a Breast Carcinoma HE slide, where a highlighted image patch shows a cluster of cells. When inspecting its neighborhood, it is seen that this cluster is not an isolated pattern, but part of a mammary lobe. \textit{Masked Context Modelling with Knowledge Distillation} aims to improve downstream performance by including context information in the feature extraction step. b) Comparison (number of parameters, FLOPs per forward pass, downstream MIL classification task AUROC) of ImageNet pre-trained feature extraction models: EfficientNetV2-L, ResNet18, and ResNet18 fine-tuned with our method using the EfficientNetV2-L as teacher. CLAM was used as MIL classification model, and performance is visualized relative to the EfficientNetV2-L model. }
  \label{fig:1}
\end{figure}



\begin{figure}[t]
    \centering
    \includegraphics[width=1.\linewidth]{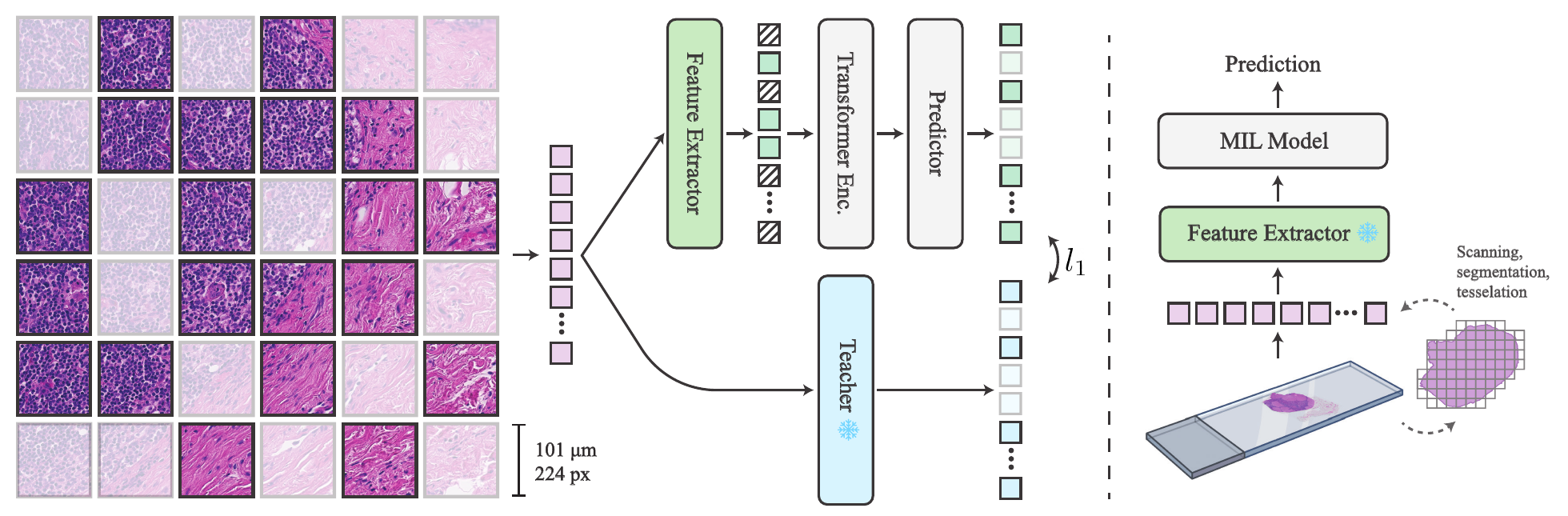}
    \caption{Proposed pipeline. During the feature extractor fine-tuning stage (left), a pre-trained feature extractor model is fed with image patches coming from a larger context window. A random subset of the patches' feature vector representations is masked, and a Transformer encoder with a predictor network is used to predict the masked instances' feature vector representations produced by a frozen teacher network, minimizing an $l_1$ loss. For the downstream task training stage (right), the Transformer and the predictor networks are discarded, and the fine-tuned feature extractor can be used in any Multiple Instance Learning pipeline.}
    \label{fig:2}
\end{figure}

\section{Introduction}
\label{sec:intro}


In Digital Pathology, specimen slides are digitised into high resolution Whole Slide Images (WSIs) of several gigapixels. This has led to the popularisation of Multiple Instance Learning (MIL) \cite{mil0,mil1} algorithms for automatic WSI classification tasks. In these algorithms, the typical pre-processing step involves tesselating the WSI into smaller image patches, or instances, and computing their feature vector representations using a pre-trained feature extractor neural network. To compute patch representations a popular choice is to use models pre-trained with supervision on ImageNet \cite{russakovsky2015imagenet}. Following this step, a MIL model is trained to produce a slide-level prediction using the complete set of instances of the WSI.

As noted in \cite{burt1987laplacian}, the common characteristic in image data is that neighboring pixels are highly correlated, and this fact extends to neighboring image patches in histology slides. A cutout of a Breast Carcinoma Hematoxylin and Eosin (HE) slide is shown in Fig. \ref{fig:1-1}), with a highlighted image patch of 220$\mathrm{\mu}$m or 224 pixels side length at $10\times$ magnification level (approximately 1$\mathrm{\mu}$m/pixel), showing a cluster of cells. HE image patches of these dimensions can capture fine-grained histology features like individual cell nucleus morphology, cell conglomerates, and small functional structures such as glands and blood vessels. These patterns tend to extend for several patches, and coarse-grained features, such as distribution and density of cell populations, arrangement of functional structures, tissue interfaces, overall tissue morphology and architecture, are revealed when examining a broader context. In a MIL pipeline, although a pre-trained feature extractor model can produce useful instance-level representations, it is the downstream MIL classifier that should detect and make sense out of the histological patterns that take more image area than a single patch.


In this work we propose to improve downstream MIL classification by fine-tuning the feature extractor model, such that representations of neighboring patches are predictive of one another. Drawing inspiration from reconstruction-based algorithms in the Self-Supervised Learning (SSL) literature \cite{vincent2008extracting,pathak2016context,devlin2018bert,radford2018improving,simmim,mae,beit}, we propose the Masked Context Modelling (MCM) task: individual image patches from a bigger context window are masked, and the feature extractor model is fine-tuned to predict the missing patches based on the visible ones.

As described above, the MCM task is challenging as there are multiple biologically plausible ways to fill the missing image patches in the masked context. Additionally, synthesizing such image patches would require an image generation model, such as a Variational Autoencoder \cite{kingma2013auto}, a Generative Adversarial Network \cite{goodfellow2020generative}, or a Diffusion Model \cite{sohl2015deep,ho2020denoising}. This image generation task is secondary, as our goal is not to produce realistically looking image patches but to learn useful representations. We propose therefore to predict instead of images, the feature vectors of a larger, pre-trained, teacher network and using them as a proxy of the visual information of the masked images. 

The contributions of this work can be summarised as follows:
\begin{itemize}
    \item We introduce the \textit{Masked Context Modelling with Knowledge Distillation (MCM+KD)} task to fine-tune the feature extractor model used in a MIL pipeline by using a larger pre-trained model as teacher. 
    \item This task can be trained even for a single epoch, between the feature extractor pre-training stage and the final training of the MIL model, improving downstream performance in two Cancer Subtype Classification tasks and a Lymph Node Metastases Detection task.
    \item We show that the fine-tuned student does not learn to copy the teacher's output. Furthermore, the student model can result in higher downstream performance than the teacher, while having fewer parameters and needing less computation to process input images (Fig. \ref{fig:1-2}).
\end{itemize}


\section{Related Work}
\label{sec:related}
\subsection{Pre-training and Fine-tuning}
It is empirically proven that transfer learning from ImageNet can improve classification performance and speed up convergence in medical image analysis applications \cite{tlm1,tlm2,tlm3}. In MIL algorithms for Digital Pathology, frozen ImageNet pre-trained models are commonly used in the feature extraction step \cite{campanella,clam,dsmil,transmil}. This is partially due to the fact that in a MIL problem, the lack of instance-level labels prevents the fine-tuning of a feature extractor model with supervision. 

A different line of research investigates how the domain shift between natural images and medical images could be alleviated by algorithms that require no labels. For example, \cite{azizi2021big} proposes to do SSL with in-domain medical data on ImageNet pre-trained models, prior to supervised fine-tuning. We refer to \cite{guan2021domain} for a comprehensive survey on Domain Adaptation in the medical image field, including unsupervised methods. 

\subsection{Reconstruction-based Algorithms}
Training models without supervision -- by masking a part of their input and teaching them to reconstruct it -- has achieved great success both in Computer Vision (CV) and Natural Language Processing (NLP). Pioneering works in CV include Denoising Autoencoders \cite{vincent2008extracting} and Context Encoders \cite{pathak2016context}. In the former, masking is applied as noise corruption, where in the latter, a region of the input image is explicitly zeroed out.

With the advent of the Transformer architecture \cite{vaswani2017attention}, Masked Language Modelling (MLM) \cite{devlin2018bert} and Causal Language Modelling \cite{radford2018improving} became the dominant SSL algorithms in NLP. MLM has inspired a plethora of SSL works in CV, yielding the Masked Image Modelling (MIM) family of algorithms \cite{beit,simmim,mae}, although these methods are restricted to Vision Transformer (ViT) architectures \cite{dosovitskiy2020image}.

\subsection{Knowledge Distillation}
Knowledge Distillation (KD) consists of training a student neural network to predict outputs produced by a teacher model. It was originally proposed by \cite{hinton2015distilling} as a model compression method, by using the prediction outputs of a larger model as soft label targets to train a smaller model. Several variants have been proposed, including the prediction of the teacher's intermediate activations, confidence maps for pose estimation \cite{fastpose}, or the use of a special token for distillation in Transformer models \cite{deit}. Notably, these model compression techniques have been applied in non-MIL image classification tasks in Digital Pathology \cite{kdhisto1,kdhisto2,kdhisto3,kdhisto4}. 

Knowledge Distillation has become increasingly popular in the SSL literature with Self-Distillation (SD) algorithms, where the pre-trained teacher network is replaced by an Exponential Moving Average (EMA) of the student \cite{byol,dino,moco,ressl}. Recently in Digital Pathology, \cite{tang2023multiple} has proposed the use of an additional SD-based loss for training supervised MIL classifiers.

\section{Methodology}
\label{sec:methodology}
We now introduce our \textit{Masked Context Modelling with Knowledge Distillation} task, designed to fine-tune the feature extractor network from the pre-processing step of a MIL pipeline. This fine-tuning step is performed before training of the WSI classification MIL model, and is independent of this model's architecture. The proposed pipeline is illustrated in Fig. \ref{fig:2}, and described below.

\textbf{Context Window}. Given a MIL pipeline where a WSI is processed as a set of square image patches of side $p$ at a specified magnification objective, we crop large square patches of side $P=s \cdot p$ at the same magnification objective. The large image patches act as context windows, and are tessellated into $s \cdot s$ square patches of side $p$. 

\textbf{Student and Teacher}. The feature extractor network $f$ to be fine-tuned acts as a student network. $f$ converts the image patches of a context window into the sequence of feature vectors \(\mathbf{x} \in \mathbb{R}^{s \cdot s \times d_{f}}\), where $d_f$ is the latent dimension of the model. Similarly, the same image patches are converted into the sequence of feature vectors \(\mathbf{y} \in \mathbb{R}^{s \cdot s \times d_{t}}\) by the network $t$, a frozen and pre-trained teacher model with latent dimension $d_t$. Feature vectors from the sequence $\mathbf{y}$ are used later as prediction targets.

\textbf{Masked Context Modelling}. Following the masking strategy of \cite{simmim,mae}, a random subset of instances from $\mathbf{x}$ is sampled using a uniform distribution, and each of the sampled instances is replaced by a learnable mask token. We add learnable positional embeddings to this sequence and feed it to a Transformer encoder model, which outputs the latent sequence representation of the masked context window, \(\mathbf{x}_L \in \mathbb{R}^{s \cdot s \times d_{f}}\).

\textbf{Predictor Network and Objective}. A predictor network predicts the feature vector sequence $\mathbf{y}$ produced by the teacher, from the latent representation of the masked context, $\mathbf{x}_L$. The whole pipeline is trained to minimize the $l_1$ loss as in SimMIM \cite{simmim}:

\begin{equation}
    L = \frac{1}{\mathrm{\Omega}(\mathbf{y}_M)} \| \mathbf{y}_M - \mathbf{y}_M' \|_{1},
\end{equation}
where \(\mathbf{y}' \in \mathbb{R}^{s \cdot s \times d_{t}}\) is the output of the predictor network, the subscript $M$ denotes the set of feature vectors that correspond to masked instances, and \(\mathrm{\Omega}(\cdot)\) is the number of elements.

Once $f$ is fine-tuned, the Transformer encoder and the predictor network are discarded. Then $f$ can be used as feature extractor network in the pre-processing step of the downstream MIL task.

\section{Experiments}
\label{sec:experiments}
\subsection{Datasets and Pre-processing}
We evaluate our method in the tasks of Breast Carcinoma Subtype Classification (BCSC), Lung Carcinoma Subtype Classification (LCSC), and Lymph Node Metastases Detection. 

\textbf{Breast Carcinoma Subtype Classification}. This dataset consists of 500 HE-stained WSIs from the TCGA-BRCA project within The Cancer Genome Atlas repository (https://www.cancer.gov/tcga). Out of the 500 slides, 351 come from Invasive Ductal Carcinoma cases, and 149 come from Invasive Lobular Carcinoma cases. Evaluation was done in a train-val-test fashion, using patient-level stratified data splits of 80\%-10\%-10\%. 

\textbf{Lung Carcinoma Subtype Classification}. This dataset consists of 500 HE-stained WSIs from the TCGA-LUAD and TCGA-LUSC projects within The Cancer Genome Atlas repository. Out of the 500 slides, 250 come from Lung Adenocarcinoma cases, and 250 come from Lung Squamous Cell Carcinoma cases. Same evaluation protocol as in the BCSC task was used. 

\textbf{Lymph Node Mestastases Detection}. This dataset is provided by the CAMELYON16 Grand Challenge (https://camelyon16.grand-challenge.org/). The training dataset comprises 270 HE-stained WSIs of sentinel lymph nodes (160 normal slides, and 110 slides containing metastases), and the test set 129 WSIs (80 normal slides, 49 containing metastases). The slides were scanned by 3DHISTECH and Hamamatsu scanners at $\times40$ magnification objective at the Radboud University Medical Center and the University Medical Center Utrecht, Netherlands. In our experiments, we divided the provided train set in 90\%-10\% stratified data splits for training and validation.

During pre-processing, the slides were tiled into large crops of 1792$\times$1792 pixels at a single magnification objective ($\times20$ for the LNMD task and $\times10$ for the BCSC and LCSC tasks), and images with less than 60\% of tissue were removed. These crops acted as context windows for the MCM+KD task, and were subsequently subdivided into image patches of 224$\times$224 pixels. In the downstream classification task, MIL models were trained considering all the 224$\times$224 image patches of a single slide as an individual sample. 

\subsection{Experimental Setup}
\label{sec:experiments:impl}

Throughout all the experiments we use an ImageNet pre-trained ResNet18 \cite{resnet} as MIL feature extractor model and ImageNet pre-trained EfficientNetV2-L \cite{effnet} as teacher model in the MCM+KD task. We take the activations before the last classification layer as feature vectors in both networks.

The Transformer encoder was parametrized with input size of 512 (the same as the ResNet18 feature vectors), 8 Multi-head Attention layers of 4 attention heads, and MLPs with hidden dimension of 3072. The predictor model is a 2-layer MLP with 1280 hidden units and output units (the same as the EfficientNetV2-L feature vectors), with a GELU \cite{gelu} hidden activation layer.

We used MCM+KD for a single epoch to fine-tune the pre-trained ResNet18 in all experiments. AdamW optimizer \cite{adamw} and batch size of 8 context windows were used. The learning rates were searched individually for each dataset, to maximize downstream CLAM \cite{clam} classification, yielding the learning rates of 0.0001, 0.001, 0.0005 for the LNMD, BCSC, and LCSC tasks, respectively, and for the other optimizer hyperparameters we kept PyTorch default values. The percentage of masked patches in a context was selected in a similar manner, masking 60\% of the patches of a context window in the LNMD task, and 50\% of the patches in a context window in the BCSC and LCSC tasks.

Downstream MIL classifiers were trained for 100 epochs, using a batch size of 1 WSI per batch, and accumulating gradients for 8 optimization steps. The models were optimized with AdamW, learning rate was searched for each dataset and classifier independently, and model selection was done based on validation loss. We report the macro-averaged AUROC as main performance metric.

The method was implemented in Python, using PyTorch \cite{pytorch} and PyTorch-Lightning \cite{lightning}. The ImageNet pre-trained weights of the feature extractor models were downloaded from Torchvision \cite{torchvision}. Implementation of SSL methods was taken from the solo-learn library \cite{solo}. Training was done on a single NVIDIA Ampere A100 GPU. The code of this project will be available in \url{https://github.com/bozeklab}.

\subsection{WSI Classification Results}
Our main experiment is the comparison of downstream MIL classification performance of a baseline ImageNet pre-trained ResNet18 used as feature-extractor model, against the same feature-extractor model, fine-tuned with MCM+KD for a single epoch. For the comparison, we include the downstream performance of MIL models that used the pre-trained EfficientNetV2-L teacher as feature extractor model. 

We compare downstream classification of three popular MIL algorithms: Attention-Based Deep Multiple Instance Learning (ABDMIL) \cite{ilse}, which popularised the use of the attention mechanism in MIL; CLAM \cite{clam}, an improvement over ABDMIL that incorporates a clustering loss in the latent space of instances; and TransMIL \cite{transmil}, a BERT-like Transformer encoder designed for Digital Pathology MIL tasks. 

The results are shown in Tables \ref{tab:res-main-brca}-\ref{tab:res-main-c16}. Our method increases the downstream performance of the ResNet18 feature extractor model across all performed comparisons, even surpassing the EfficientNetV2-L teacher model in some scenarios.

\begin{table}[h]
    \centering
    \caption{ Breast Carcinoma Subtype Classification results.}\label{tab:res-main-brca}
    \begin{tabular}{l c c c c c}
        \toprule
        Feature extraction              & & ABDMIL      & CLAM        & TransMIL \\
        \midrule
        ResNet18                        & &        0.75 &        0.82 &         0.85 \\
        ResNet18 (MCM+KD)       & & \bftab 0.76 & \bftab 0.92 & \bftab 0.89 \\[5pt]
        EfficientNetV2-L (teacher)      & &        0.72 &        0.86 &         0.94 \\
        \bottomrule
    \end{tabular}
\end{table}

\begin{table}[h]
    \centering
    \caption{ Lung Carcinoma Subtype Classification results.}\label{tab:res-main-lung}
    \begin{tabular}{l c c c c c}
        \toprule
        Feature extraction               & & ABDMIL      & CLAM        & TransMIL \\
        \midrule
        ResNet18                         & &        0.83 &        0.86 &        0.68 \\
        ResNet18 (MCM+KD)        & & \bftab 0.91 & \bftab 0.90 & \bftab 0.77 \\[5pt]
        EfficientNetV2-L (teacher)       & &        0.82 &        0.91 &        0.82\\
        \bottomrule
    \end{tabular}
\end{table}

\begin{table}[h]
    \centering
    \caption{ Lymph Node Metastases Detection results.}\label{tab:res-main-c16}
    \begin{tabular}{l c c c c c}
        \toprule
        Feature extraction              & & ABDMIL      & CLAM        & TransMIL \\
        \midrule
        ResNet18                        & &        0.58 &        0.79 &        0.69  \\
        ResNet18 (MCM+KD)       & & \bftab 0.69 & \bftab 0.84 & \bftab 0.76  \\[5pt]
        EfficientNetV2-L (teacher)      & &        0.76 &        0.75 &        0.76  \\
        \bottomrule
    \end{tabular}
\end{table}

\subsection{Self-Distillation Scenario}
As suggested in \cite{azizi2021big}, a possible solution to the domain shift problem when using ImageNet pre-trained models in medical image analysis is the use of SSL algorithms with in-domain data. In this experiment we use self-distillation where the pre-trained teacher model is replaced by an Exponential Moving Average (EMA) of the student network.

We compare this Self-Distillation version of the MCM task (denoted \textit{MCM+SD}) against other SSL algorithms that make use of an EMA teacher: BYOL \cite{byol}, DINO \cite{dino}, MOCOv3 \cite{moco}, and ReSSL \cite{ressl}. We include SimCLR \cite{simclr} and SimSiam \cite{simsiam} for comparison as well, although these methods are not from the Self-Distillation family.

This experiment was evaluated using CLAM downstream classification, and hyperparameters were kept the same as in the previous experiment. For the Self-Distillations algorithms, including MCM+SD, we use an EMA $\tau$ of 0.999.

The results in Table \ref{tab:res-sd} show that our MCM+SD method was the only one improving the results over the baseline pre-trained ResNet18 for the BCSC and LCSC tasks.  However, none of the tested methods improved LNMD classification. Although the compared methods are well established for self-supervised pre-training, our results highlight the role of the pre-trained teacher for fine-tuning a feature extractor model with the limited compute budget of a single epoch.

\begin{table}[h]
    \centering
    \caption{Comparison of downstream MIL classification using CLAM when doing pre-processing fine-tuning with different SSL methods, and a Self-Distillation version of the Masked Context Modelling task.}\label{tab:res-sd}
    \begin{tabular}{l c c c}
        \toprule
        Method & BCSC & LCSC & LNMD \\
        \midrule
        Baseline                        & 0.82 & 0.86 & \bftab 0.79 \\
        BYOL            & 0.70 & 0.66 & 0.62 \\
        DINO            & 0.50 & 0.54 & 0.60 \\
        MOCOv3          & 0.82 & 0.82 & 0.67 \\
        ReSSL           & 0.62 & 0.71 & 0.50 \\
        SimCLR          & 0.79 & 0.84 & 0.65 \\
        SimSiam         & 0.51 & 0.47 & 0.61 \\             
        MCM+SD                   & \bftab 0.88 & \bftab 0.89 & 0.65 \\[5pt]
        MCM+KD                    & 0.92 & 0.90 & 0.84 \\
        \bottomrule
    \end{tabular}
\end{table}

\subsection{Ablation Study} 
\label{ablation}
Here we perform an ablation study to analyze individual components of MCM+KD (Table \ref{tab:res-ab}). In our experiment denoted as \textit{MCM}, we fine-tune the feature extractor model by predicting the pixels of the masked image patches from the context window, omitting the knowledge distillation task. We also investigate the need of the MCM task: we omit the masking of patches and the Transformer encoder, and use a simple Knowledge Distillation (KD) approach instead. The KD consists of directly predicting the feature vectors of the teacher model. 

Finally, we do an experiment (denoted by \textit{CM + KD}) where the context modelling Transformer, predictor network, and teacher network are kept, but the patches are not masked. This test was performed to verify the need of such design choice. In contrast to an image reconstruction task, in the knowledge distillation scenario our model is not at risk of learning the identity function or memorizing the input data.

The ablation study was done with a CLAM downstream classifier, keeping the previous hyperparameters. For the MCM experiment, the MLP predictor was replaced by a single linear layer, as an MLP as described in \ref{sec:experiments:impl} with $224\times224\times3$ hidden units is infeasible. Our results show that only our complete pipeline can improve the downstream CLAM performance over the baseline pre-trained ResNet18 feature extractor.

\begin{table}[h]
    \centering
    \caption{Ablation study results.}\label{tab:res-ab}
    \begin{tabular}{l c c}
        \toprule
        Model & LNMD & LCSC \\
        \midrule
        Baseline & 0.79 & 0.86 \\
        MCM & 0.42 & 0.72\\
        KD & 0.76 & 0.86\\ 
        CM + KD & 0.56 & 0.82 \\ 
        MCM + KD & \bftab 0.84 & \bftab 0.90\\
        
        \bottomrule
    \end{tabular}
\end{table}

\subsection{Qualitative Results}
The aim of this experiment is to visually evaluate the effect that our method has on the feature vectors of the ResNet18 feature extractor, independent of the downstream MIL classifier. We take slides from the CAMELYON16 test split which contain annotated metastases regions. We select a patch within a metastasis region and compute the cosine similarity of its feature vector and the feature vectors of the rest of the patches in the slide. We show these cosine similarity values as heatmap visualizations. We do this procedure with the feature vectors obtained by the baseline ResNet18, the ResNet18 fine-tuned with MCM+KD, and the EfficientNetV2-L teacher.

\begin{figure}[t!]
    \centering
    \includegraphics[width=1.\linewidth]{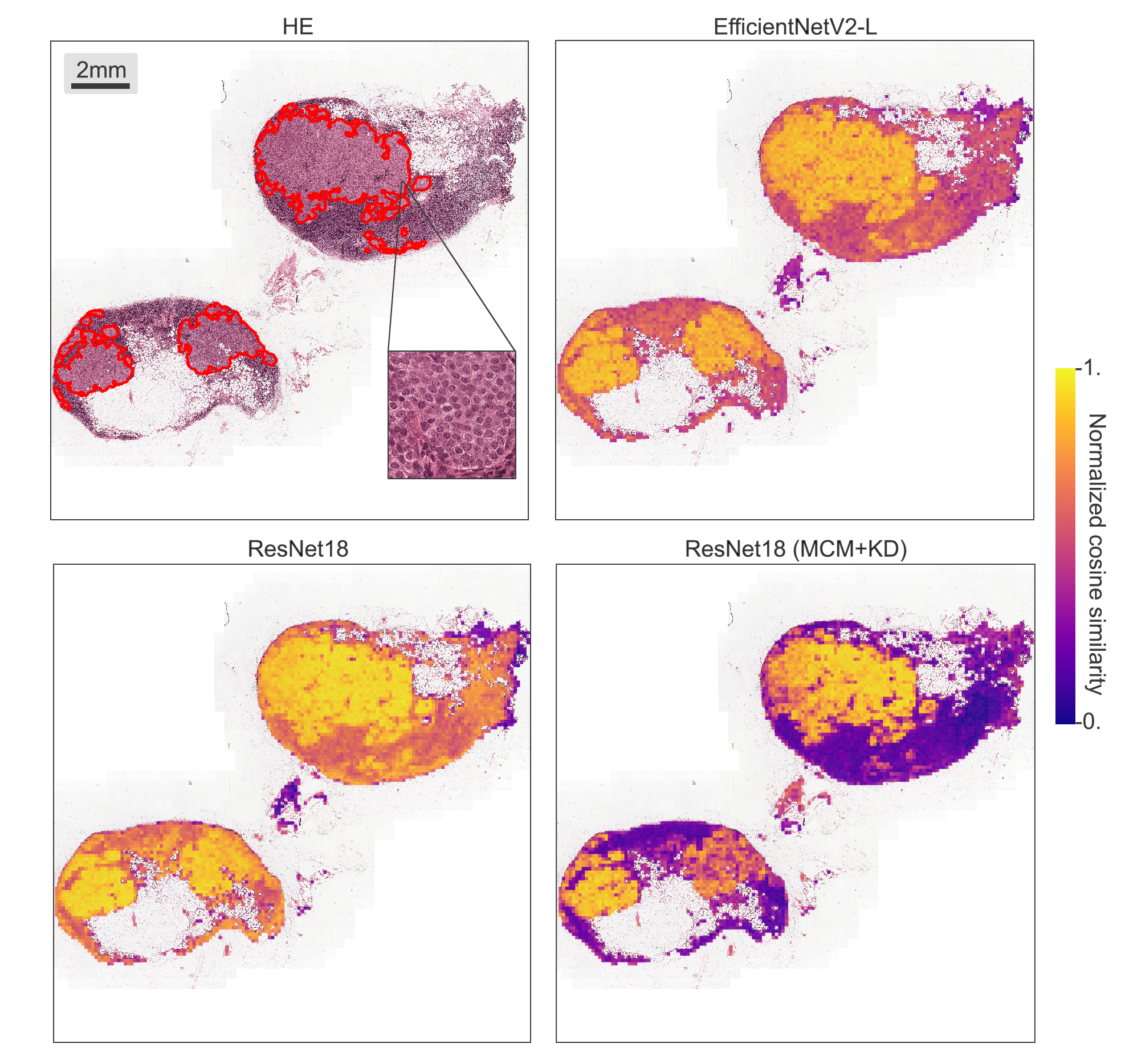}
    \caption{Heatmap visualizations of the cosine similarity between the feature vector of a patch from a metastasis region and the rest of the feature vectors of the slide.}
    \label{fig:heatmaps}
\end{figure}

Fig. \ref{fig:heatmaps} shows heatmaps of a WSI with metastases. These heatmaps depict that our method decreases the similarity between feature vectors of metastases and feature vectors of normal tissue. The better visual discrimination between biologically different tissue regions in the heatmaps of the baseline and our fine-tuned ResNet18 models is consistent with the performance increase in the downstream MIL tissue classification task. The difference in the heatmaps of the teacher and the student models is consistent with the results in Sec. \ref{ablation} showing that downstream performance is not achieved by copying the teacher's output.

\section{Discussion and Conclusion}
In this paper we demonstrate how downstream MIL classification can be improved by fine-tuning the feature extractor model using our proposed Masked Context Modelling with Knowledge Distillation task. This algorithm can be trained for a single epoch and is agnostic of the MIL model used in the downstream task. All our experiments were done using an EfficientNetV2-L teacher to fine-tune a ResNet18 feature extractor, with both networks being pre-trained with supervision on ImageNet. 

Our results show that across the datasets and MIL models, our method improves the downstream performance. It is worth noting that although the teacher model is considerably bigger, computationally more expensive, and outperforms the ResNet18 in the original ImageNet classification pre-training task, the feature vectors it produces do not always result in better downstream performance. Nevertheless, the teacher model is still effective for distilling its knowledge to the student.

Using the same experimental setup, we fine-tuned the same ResNet18 with different SSL algorithms. These training scenarios did not however improve the results over the baseline ImageNet-pre-trained ResNet18. Our self-distillation version of the MCM task was not successful in the LNMD dataset, but it did improve the results in the BCSC and LCSC tasks, although not matching the results of MCM+KD with an EfficientNetV2-L teacher. This experiment suggests that even though SSL algorithms and self-distillation are effective in a full pre-training scenario, knowledge distillation from a larger pre-trained teacher is very useful in a single epoch of fine-tuning.

In our ablation study, we show that the individual components of our method do not work in a standalone setting, and the success of our method comes from the proper combination of these algorithms and principles. Our results show that the performance of our method is not explained solely by knowledge distillation, and we verify this by visual examination of cosine similarity heatmaps. The prediction of masked instances plays an essential role in our algorithm, and we leave exploring more sophisticated masking policies, such as adversarial masking \cite{adios}, for future work.

Altogether, we indicate the importance of context in achieving the best representations of WSI patches. Our masking and KD approach allows to efficiently encode this context and to make use of it in the further WSI analysis tasks.

\textbf{Acknowledgements.}
We thank France Rose and Piotr Wójcik for the fruitful discussions and feedback. Both K.B. and J.I.P. were hosted by the Center for Molecular Medicine Cologne throughout this research. K.B. and J.I.P. were supported by the BMBF program Junior Group Consortia in Systems Medicine (01ZX1917B) and BMBF program for Female Junior Researchers in Artificial Intelligence (01IS20054).
%
%
\bibliographystyle{splncs04}
\bibliography{egbib}
\end{document}